\pdfoutput=1
\documentclass[11pt]{article}
\usepackage[round]{natbib}
\usepackage[]{ACL2023}
\usepackage{times}
\usepackage{latexsym}
\usepackage{float}
\usepackage[T1]{fontenc}
\usepackage[utf8]{inputenc}
\usepackage{microtype}
\usepackage{inconsolata}
\usepackage{graphicx}
\usepackage{tikz}
\usepackage{fontawesome5}
\usepackage{tabularx}
\usepackage{xcolor}
\usepackage{mathtools}
\usetikzlibrary{shapes,arrows,positioning,fit,backgrounds,calc}
\usepackage{booktabs} 
\usepackage{multirow}
\usepackage{subcaption}
\usepackage{CJKutf8}
\usepackage{amssymb}
\usepackage{enumitem}

\title{Token Masking Improves Transformer-Based Text Classification}

\author{
  Xianglong Xu \hspace{1em} John Bowen \hspace{1em} Rojin Taheri \\
  University of Pittsburgh \\
  School of Computing and Information \\
  \texttt{\{xix110, jeb386, rot64\}@pitt.edu}
}

\begin{document}
\maketitle
\begin{abstract}
While transformer-based models achieve strong performance on text classification, we explore whether masking input tokens can further enhance their effectiveness. We propose token masking regularization, a simple yet theoretically motivated method that randomly replaces input tokens with a special \texttt{[MASK]} token at probability $p$. This introduces stochastic perturbations during training, leading to implicit gradient averaging that encourages the model to capture deeper inter-token dependencies. Experiments on language identification and sentiment analysis—across diverse models (mBERT, Qwen2.5-0.5B, TinyLlama-1.1B)—show consistent improvements over standard regularization techniques. We identify task-specific optimal masking rates, with $p=0.1$ as a strong general default. We attribute the gains to two key effects: (1) input perturbation reduces overfitting, and (2) gradient-level smoothing acts as implicit ensembling.
\end{abstract}

\section{Introduction}
\label{sec:Introduction}
Attention mechanisms have transformed NLP by capturing long-range dependencies and contextual relationships. However, token-level lexical information can act as noise, leading attention heads to overfit to superficial cues rather than learning generalizable patterns.

As shown in \citet{nitish2014dropout}, one effective regularization approach is to average over an ensemble of model configurations, approximated via Dropout. Inspired by this, we hypothesize that attention heads can be regularized by averaging over all plausible attention matrices, mitigating token-level noise and encouraging robustness.

Instead of modifying model architecture, our approach perturbs the input data. By masking tokens with a certain probability, we simulate an ensemble over data configurations that share the same underlying structure. This encourages attention heads to focus on invariant relational patterns, providing a complementary alternative to model-level regularization like Dropout.

Our method is motivated by the structured nature of code-switching, where speakers alternate between languages within a sentence \cite{article}. Prior studies show such switches follow context-sensitive patterns \cite{DAVIES1996148}. For example, "\begin{CJK}{UTF8}{gbsn}我喜欢学习\end{CJK} NLP" is natural, while "\begin{CJK}{UTF8}{gbsn}我\end{CJK} like \begin{CJK}{UTF8}{gbsn}学习自然语言处理\end{CJK}"—constructed for illustration—is less grammatical. In a sentence like "X means hello in English", the model need not know the language of "X", only that it is distinct. This highlights the need to model linguistic boundaries and roles beyond lexical identity.

We thus propose a simple yet effective data-level regularization method—token masking regularization—that perturbs training inputs to promote attention to deeper patterns, particularly in code-switched text.

\section{Related Work}
\label{sec:Related Work}
Dropout introduces noise to neural network units, interpreted as implicit model averaging over submodels \cite{nitish2014dropout}. Similarly, token masking injects input-level noise, akin to perturbing hidden units \cite{10.5555/1756006.1953039} or marginalized corrupted features (MCF) \cite{pmlr-v28-vandermaaten13}. Our approach replaces tokens with a default value without modifying the loss function, inducing gradient-level "averaging" that dampens reliance on specific tokens and encourages robustness to lexical variation.

The masked language modeling (MLM) objective in BERT \cite{devlin2019bertpretrainingdeepbidirectional} popularized token masking for pretraining, replacing input tokens with [MASK] and training models to recover original tokens. This fosters contextual representations grounded in bidirectional token relationships.

MaskLID \cite{kargaran2024masklid} addresses language identification in code-switching by iteratively masking dominant-language (L1) tokens, encouraging the model to detect secondary (L2) languages—unlike traditional models that struggle with mixed-language input.

\section{Method}
\label{sec:Method}
We propose training Transformers with probabilistic token masking: during each forward pass, every input token is replaced with \texttt{[MASK]} with probability $p$. This stochastic corruption disrupts surface lexical cues, forcing the model to rely on contextual dependencies rather than memorizing token-level patterns. The model is trained end-to-end on the target task using these stochastically masked sequences, learning to leverage sparse, incomplete token signals to build robust task-specific representations.  

This stochastic masking introduces structured noise into the gradient flow. When averaged over multiple corruption configurations, the gradients converge to a smoothed update direction, effectively approximating an ensemble of masked sub-networks.

Formally, given input sequence $\mathbf{X} \in \mathbb{R}^{T \times d}$, we replace each token $\mathbf{x}_t$ with mask token $\mathbf{x}_{\text{mask}}$ with probability $p$, producing corrupted sequence $\mathbf{X}'$. Let $\mathcal{M} \subseteq \{1,\ldots,T\}$ denote masked positions.

For self-attention parameters $\mathbf{W}_Q$ (similarly $\mathbf{W}_K$, \allowbreak $\mathbf{W}_V$), define the gradient difference between masked and original inputs:

{\small
\begin{align}
\Delta\left(\frac{\partial\mathcal{L}}{\partial \mathbf{W}_Q}\right) 
&= \mathbf{X}'^\top\frac{\partial\mathcal{L}'}{\partial \mathbf{Q}'} - \mathbf{X}^\top\frac{\partial\mathcal{L}}{\partial \mathbf{Q}} \nonumber \\
&= \underbrace{\sum_{t \in \mathcal{M}} (\mathbf{x}_{\text{mask}} - \mathbf{x}_t)^\top \frac{\partial\mathcal{L}'}{\partial \mathbf{q}'_t}}_{\text{Input Corruption}} 
+ \underbrace{\mathbf{X}^\top\left(\frac{\partial\mathcal{L}'}{\partial \mathbf{Q}'} - \frac{\partial\mathcal{L}}{\partial \mathbf{Q}}\right)}_{\text{Update Coupling}}
\end{align}
}

This generalizes to all parameters $\theta \in \{\mathbf{W}_Q, \mathbf{W}_K, \mathbf{W}_V\}$:

{\small
\begin{equation}
\Delta\left(\frac{\partial\mathcal{L}}{\partial \theta}\right) = 
\sum_{t \in \mathcal{M}} (\mathbf{x}_{\text{mask}} - \mathbf{x}_t)^\top \frac{\partial\mathcal{L}'}{\partial \theta'_t} 
+ \mathbf{X}^\top\left(\frac{\partial\mathcal{L}'}{\partial \theta'} - \frac{\partial\mathcal{L}}{\partial \theta}\right)
\end{equation}
}

Taking expectation over random masking configurations $\mathcal{M}$ with independent probability $p$:

{\small
\begin{align}
\mathbb{E}_{\mathcal{M}}[\Delta] 
&= \underbrace{p\sum_{t=1}^T (\mathbf{x}_{\text{mask}} - \mathbf{x}_t)^\top 
\mathbb{E}\left[\frac{\partial\mathcal{L}'}{\partial \theta'_t} \big| t \in \mathcal{M}\right]}_{\text{Explicit Masking Term}} \notag \\
&\quad + \underbrace{\mathbb{E}_{\mathcal{M}}\left[\mathbf{X}^\top\left(\frac{\partial\mathcal{L}'}{\partial \theta'} - \frac{\partial\mathcal{L}}{\partial \theta}\right)\right]}_{\text{Implicit Coupling Term}}
\end{align}
}

Under the robustness assumption $\mathbb{E}[\partial\mathcal{L}'/\partial\theta'] \approx \mathbb{E}[\partial\mathcal{L}/\partial\theta]$, the coupling term vanishes. The remaining term reveals gradient averaging:

{\small
\begin{align}
\mathbb{E}\left[\frac{\partial\mathcal{L}'}{\partial\theta_t}\right] &= p\frac{\partial\mathcal{L}_{\text{mask}}}{\partial\theta_t} + (1-p)\frac{\partial\mathcal{L}_{\text{original}}}{\partial\theta_t} \\
\mathbb{E}[\Delta] &\propto \sum_{t=1}^T \Big[p^2\mathbf{x}_{\text{mask}}^\top\frac{\partial\mathcal{L}_{\text{mask}}}{\partial\theta_t} + p(1-p)\mathbf{x}_t^\top\frac{\partial\mathcal{L}_{\text{original}}}{\partial\theta_t}\Big]
\end{align}
}

Sustained masking therefore implements parameter updates that converge to:

{\small
\begin{equation}
\mathbb{E}[\nabla_\theta \mathcal{L}] \approx \frac{1}{|\mathcal{C}|} \sum_{c\in\mathcal{C}} \nabla_\theta \mathcal{L}_c
\end{equation}
}

Let $\mathcal{C}$ be the set of all masking configurations under Bernoulli probability $p$. Each configuration $c \in \mathcal{C}$ defines a unique masked input $\mathbf{X}^{(c)}$ and resulting loss $\mathcal{L}_c$. Then:

\[
\mathbb{E}[\nabla_\theta \mathcal{L}] = \sum_{c \in \mathcal{C}} P(c)\, \nabla_\theta \mathcal{L}_c
\]

That is, training under token masking minimizes the expected loss over the distribution of masked inputs, and the corresponding parameter updates are gradients averaged over stochastic perturbations of the input. This forms a data-dependent ensemble in gradient space, similar in spirit to Dropout, but operating on token-level input perturbations instead of internal activations.

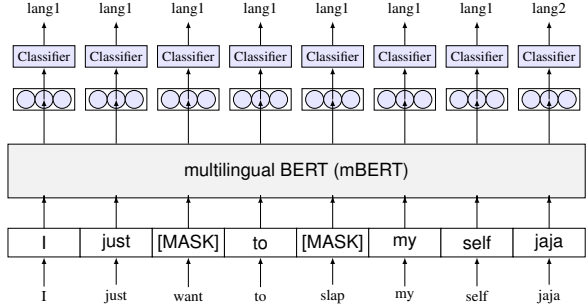
\begin{figure}[h!]
    \centering
    \resizebox{1 \linewidth}{!}{
    \begin{tikzpicture}[
        box/.style={draw, rectangle, minimum width=2cm, minimum height=0.8cm, fill=white, thick},
        mbert/.style={draw, rectangle, minimum width=16cm, minimum height=1.5cm, fill=gray!10, thick},
        classifier/.style={draw, rectangle, minimum width=1.5cm, minimum height=0.6cm, fill=blue!10, thick, font=\normalsize},
        neuron/.style={draw, circle, minimum size=0.5cm, fill=blue!10, thick},
        neuron_frame/.style={draw, rectangle, minimum width=1.7cm, minimum height=0.6cm, thick, thick},
        arrow/.style={->, >=latex, thick},
        every node/.style={font=\sffamily\Large},
        label/.style={font=\large}
    ]

        \node[label] (l1) at (-7,5.5) {lang1};
        \node[label] (l2) at (-5,5.5) {lang1};
        \node[label] (l3) at (-3,5.5) {lang1};
        \node[label] (l4) at (-1,5.5) {lang1};
        \node[label] (l5) at (1,5.5) {lang1};
        \node[label] (l6) at (3,5.5) {lang1};
        \node[label] (l7) at (5,5.5) {lang1};
        \node[label] (l8) at (7,5.5) {lang2};

        \foreach \x in {-7,-5,-3,-1,1,3,5,7} {
            \node[neuron_frame] at (\x,3) {};
            \node[neuron] (n1\x) at (\x-0.5,3) {};
            \node[neuron] (n2\x) at (\x,3) {};
            \node[neuron] (n3\x) at (\x+0.5,3) {};
            \node[classifier] (clf\x) at (\x,4.2) {Classifier};
        }

        \node[mbert] (mbert) at (0,1) {multilingual BERT (mBERT)};

        \node[box] (t1) at (-7,-1) {I};
        \node[box] (t2) at (-5,-1) {just};
        \node[box] (t3) at (-3,-1) {[MASK]};
        \node[box] (t4) at (-1,-1) {to};
        \node[box] (t5) at (1,-1) {[MASK]};
        \node[box] (t6) at (3,-1) {my};
        \node[box] (t7) at (5,-1) {self};
        \node[box] (t8) at (7,-1) {jaja};

        \node[label] (o1) at (-7,-2.5) {I};
        \node[label] (o2) at (-5,-2.5) {just};
        \node[label] (o3) at (-3,-2.5) {want};
        \node[label] (o4) at (-1,-2.5) {to};
        \node[label] (o5) at (1,-2.5) {slap};
        \node[label] (o6) at (3,-2.5) {my};
        \node[label] (o7) at (5,-2.5) {self};
        \node[label] (o8) at (7,-2.5) {jaja};

        \foreach \x in {-7,-5,-3,-1,1,3,5,7} {
            \draw[arrow] (mbert.north -| \x,1) -- (\x,3);
        }

        \foreach \x in {-7,-5,-3,-1,1,3,5,7} {
            \draw[arrow] (\x,3) -- (clf\x.south);
        }

        \foreach \x/\l in {-7/l1, -5/l2, -3/l3, -1/l4, 1/l5, 3/l6, 5/l7, 7/l8} {
            \draw[arrow] (clf\x.north) -- (\l.south);
        }

        \foreach \i in {-7,-5,-3,-1,1,3,5,7} {
            \draw[arrow, shorten >=20pt] ({\i},-0.6) .. controls ({\i},{0.2}) and ({\i},{0.6}) .. ({\i},1);
        }

        \foreach \i/\j in {t1/o1, t2/o2, t3/o3, t4/o4, t5/o5, t6/o6, t7/o7, t8/o8} {
            \draw[arrow] (\j) -- (\i);
        }


    \end{tikzpicture}
    }
    \caption{Visualization of mBERT's implementation (sentence number 28, from the (SPA-ENG) training dataset).}
    \label{fig1:mask_out}
\end{figure}

Figure~\ref{fig1:mask_out} demonstrates this via SPA-ENG sample 28, where masked tokens compel sophisticated code-switch detection through residual syntactic/semantic cues rather than lexical memorization.  

\section{Experiments}
\label{sec:Experiments}
\subsection{Datasets and Data Preparation}
\label{sec:Datasets and Data Preparation}
We leverage the Linguistic Code-switching Evaluation (LinCE) benchmark \cite{aguilar2020lince}, which comprises ten corpora across four code-switched language pairs. We focus on two key tasks: 

\textbf{Language Identification (LID)}, a token-level sequence labeling task for Spanish-English (SPA-ENG) and Nepali-English (NEP-ENG), where each token is annotated as Lang1 (English), Lang2 (Spanish/Nepali), or "Other"—with non-linguistic tokens (e.g., emoticons, URLs) mapped to "Other" following \cite{molina2019overview}; 

To evaluate cross-lingual generalization, we adopt a zero-shot transfer setup: models are trained on SPA-ENG data and evaluated on NEP-ENG without adaptation, testing their ability to capture language-agnostic code-switching patterns.

\textbf{Sentiment Analysis (SA)}, a sentence-level classification task assigning positive, negative, or neutral labels to SPA-ENG utterances.

\subsection{Models}
\label{sec:Models}
We benchmark three transformer-based models: \textbf{Multilingual BERT (mBERT)}, chosen for its strong cross-lingual transfer baseline \cite{pires2019multilingualmultilingualbert}, \textbf{Qwen2.5-0.5B} \cite{qwen2025qwen25technicalreport}, and \textbf{TinyLlama-1.1B} \cite{zhang2024tinyllamaopensourcesmalllanguage}.

\subsection{Class Balancing}
\label{sec:Class Balancing}
To address label imbalance, we compute adaptive class weights using an inverse-frequency scheme with moderated scaling. For each class $c$ containing $N_c$ samples, the weight $w_c$ is derived as:
$$
w_c = \frac{\sqrt{\frac{N_{\text{total}}}{N_{\text{classes}} \cdot N_{\text{c}}}}}{\sum\limits_{k=1}^{N_{\text{classes}}}\sqrt{\frac{N_{\text{total}}}{N_{\text{classes}} \cdot N_k}}} \cdot N_{\text{classes}}
$$
where $N_{\text{total}} = \sum_{c=1}^{N_{\text{classes}}} N_c$ (total samples),  $N_c$ = samples in class $c$ and $N_{\text{classes}}$ = total number of classes. 

\section{Results}
\label{sec:Results}
We evaluate token masking effectiveness using weighted F1 scores across LinCE benchmarks: Spanish-English (LID/SA) and Nepali-English (cross-lingual LID). Systematic testing of masking probabilities \( p \in [0.0, 0.5] \) reveals three critical findings (Tables~\ref{tab:LID(SPA-ENG)}--\ref{tab:SA(SPA-ENG)}). 

First, we observe an \textbf{optimal masking threshold}: most configurations achieve peak performance at \( p = 0.1 \), with performance degrading beyond \( p = 0.3 \). Notable exceptions include Qwen2.5-0.5B (LID, \( p = 0.3 \)) and mBERT (SA, \( p = 0.5 \)). 

Second, the results suggest \textbf{capacity-driven robustness}, as model size correlates with masking tolerance. For instance, TinyLlama-1.1B achieves peak NEP-ENG LID performance at \( p = 0.3 \) (F1: 0.8320), outperforming both Qwen2.5-0.5B (F1: 0.8233 at \( p = 0.5 \)) and mBERT (F1: 0.7585 at \( p = 0.1 \)). 

Third, we find \textbf{task-sensitivity} in masking: LID consistently benefits from controlled masking with model size roughly proportional to optimal \( p \), while SA shows greater variance, with best performance generally concentrated at \( p = 0.1 \).  

These results demonstrate token masking's dual mechanism: (1) regularization against surface pattern overfitting by suppressing token-specific biases, and (2) preservation of structural linguistic signals through partial context availability. The task-dependent optimal \( p \) reflects the tension between these effects---higher capacity models can leverage richer contextual cues to offset information loss, particularly in cross-lingual settings.  

\begin{table*}[h!]
    \centering
    \footnotesize
    \begin{tabular}{lcccccc}
        \toprule
        \textbf{Transformer-based Models} & \multicolumn{6}{c}{\textbf{F1(Weighted) Score at Different $\lambda$ Masking out probability}} \\
        \cmidrule(lr){2-7}
        & 0.0 & 0.1 & 0.2 & 0.3 & 0.4 & 0.5 \\
        \midrule
        mBERT & 0.9692 & \textbf{0.9706} & 0.9705 & 0.9699 & 0.9669 & 0.9652 \\
        Qwen2.5-0.5B & 0.9356 & 0.9393 & 0.9392 & \textbf{0.9394} & 0.9383 & 0.9349 \\
        TinyLlama-1.1B & 0.9497 & \textbf{0.9517} & \textbf{0.9517} & 0.9502 & 0.9475 & 0.9453 \\
        \bottomrule
    \end{tabular}
    \caption{Language Identification (LID) performance on Spanish-English code-switching (SPA-ENG).}
    \label{tab:LID(SPA-ENG)}
\end{table*}

\begin{table*}[h!]
    \centering
    \footnotesize
    \begin{tabular}{lcccccc}
        \toprule
        \textbf{Transformer-based Models} & \multicolumn{6}{c}{\textbf{F1(Weighted) Score at Different $\lambda$ Masking out probability}} \\
        \cmidrule(lr){2-7}
        & 0.0 & 0.1 & 0.2 & 0.3 & 0.4 & 0.5 \\
        \midrule
        mBERT & 0.7385 & \textbf{0.7585} & 0.7468 & 0.7325 & 0.6966 & 0.6772 \\
        Qwen2.5-0.5B & 0.8186 & 0.8188 & 0.8157 & 0.8232 & 0.8175 & \textbf{0.8233} \\
        TinyLlama-1.1B & 0.8024 & 0.8239 & 0.8276 & \textbf{0.8320} & 0.8270 & 0.8256 \\
        \bottomrule
    \end{tabular}
    \caption{Cross-lingual LID performance on unseen Nepali-English data (NEP-ENG).}
    \label{tab:LID(NEP-ENG)}
\end{table*}

\begin{table*}[h!]
    \centering
    \footnotesize
    \begin{tabular}{lcccccc}
        \toprule
        \textbf{Transformer-based Models} & \multicolumn{6}{c}{\textbf{F1(Weighted) Score at Different $\lambda$ Masking out probability}} \\
        \cmidrule(lr){2-7}
        & 0.0 & 0.1 & 0.2 & 0.3 & 0.4 & 0.5 \\
        \midrule
        mBERT & 0.5104 & 0.5273 & 0.5264 & 0.5179 & 0.5284 & \textbf{0.5325} \\
        Qwen2.5-0.5B & 0.4918 & \textbf{0.5399} & 0.5198 & 0.5122 & 0.4859 & 0.4565 \\
        TinyLlama-1.1B & 0.5197 & \textbf{0.5317} & 0.5270 & 0.5033 & 0.4623 & 0.4313 \\
        \bottomrule
    \end{tabular}
    \caption{Sentiment Analysis (SA) performance on Spanish-English code-switching (SPA-ENG).}
    \label{tab:SA(SPA-ENG)}
\end{table*}

\section{Discussion}
\label{sec:Discussion}
\begin{figure}[h]
    \centering
    \includegraphics[width=0.45\textwidth]{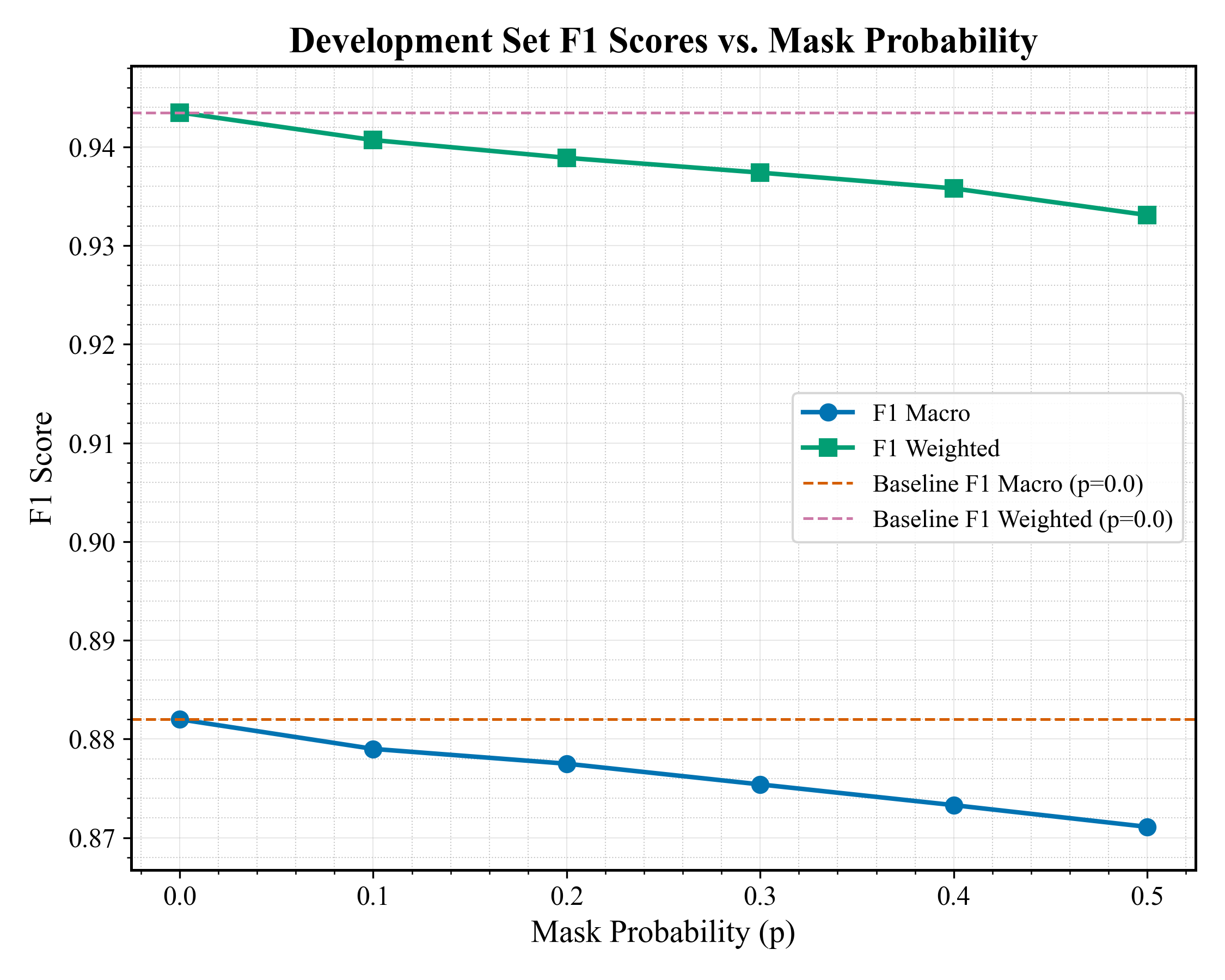}
    \caption{F1 vs. mask-out probability (POS)}
    \label{fig:f1_scores_vs_mask_probability}
\end{figure}

Our initial exploration considered POS tagging and NER but excluded them due to fundamental incompatibilities: (1)\textbf{NER Architecture Mismatch:} Requires span prediction capabilities beyond standard Transformer token classification layers, and (2)\textbf{POS Sensitivity:} Exhibits monotonic performance degradation with increasing masking (Fig.~\ref{fig:f1_scores_vs_mask_probability}), suggesting moderate label granularity (17 tags) limits regularization benefits.

This inverse correlation between POS accuracy and masking intensity highlights a critical constraint: the effectiveness of token masking diminishes as task label granularity increases. Nevertheless, consistent improvements across LID and SA tasks on multiple datasets demonstrate the robustness of our approach. Future work will explore adaptations of the method to broader or more diverse task settings.

\begin{figure}[h]
    \centering
    \includegraphics[width=0.45\textwidth]{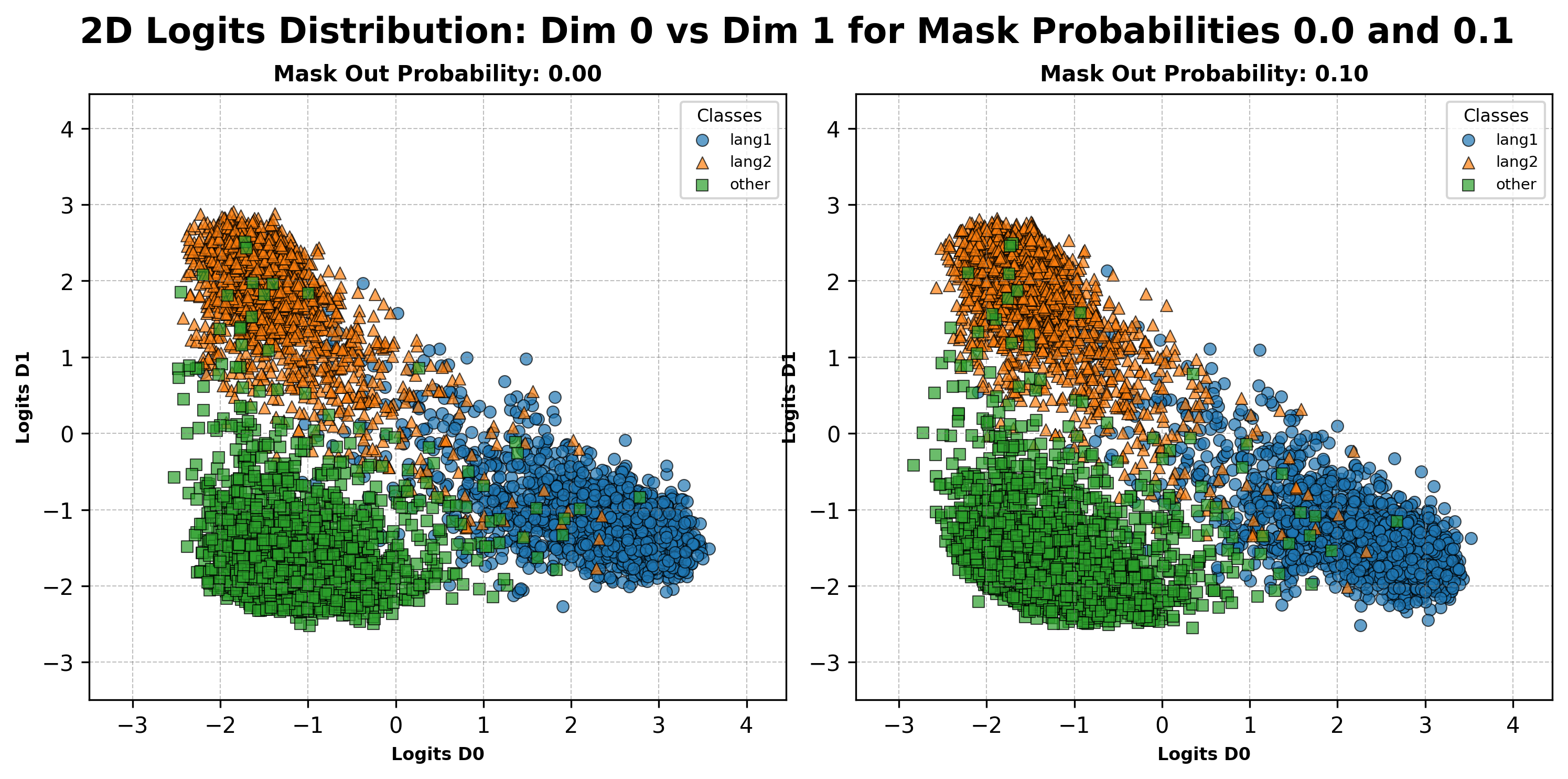}
    \caption{2D projections of logits from the LID task under different mask-out probabilities.}
    \label{fig:logits_distribution}
\end{figure}

We visualize mBERT logits from language identification under mask probabilities $p=0$ and $p=0.1$ (Fig.~\ref{fig:logits_distribution}). With the introduction of 10\% random token masking, the model's logits distribution in the feature space exhibits mild dispersion, while the cluster boundaries of core categories remain stable. This moderate distribution perturbation aligns with the observed F1-score improvement, indicating that masking implicitly regularizes the model to mitigate overfitting and encourages reliance on more robust contextual features, thereby enhancing the generalizability of classification decisions. 

\section{Conclusion} 
\label{sec:Conclusion}
We propose a simple yet effective token-level masking strategy applied during training to enhance Transformer-based classification. Extensive experiments across multiple tasks and model sizes identify \( p = 0.1 \) as a robust default masking rate. Our findings suggest that this approach encourages deeper semantic representations by reducing reliance on surface token co-occurrences. Future work includes exploring adaptive masking schedules and extending the method to encoder-decoder architectures, with the goal of supporting broader cross-lingual transfer and generative tasks.

\section*{Acknowledgements}  
We would like to express our sincere gratitude to Prof. Michael Miller Yoder and PhD student Zhuochun Li from the University of Pittsburgh for their valuable guidance, constructive feedback, and continuous support throughout this work. Their insights greatly contributed to the development and refinement of our methodology and experiments.

\bibliography{references}

\end{document}